\documentclass[asi,article,accept,pdftex,moreauthors]{Definitions/mdpi} 
\firstpage{1} 
\pubvolume{1}
\issuenum{1}
\articlenumber{0}
\pubyear{2022}
\copyrightyear{2022}
\datereceived{} 
\dateaccepted{} 
\datepublished{} 
\hreflink{https://doi.org/} 

\Title{Big Data for Traffic Estimation and Prediction: A Survey of Data and Tools}

\TitleCitation{Big Data for Traffic Estimation and Prediction: A Survey of Data and Tools}

\Author{Weiwei Jiang $^{1,}$*\orcidA{} and Jiayun Luo $^{2}$}

\AuthorNames{Weiwei Jiang and Jiayun Luo}

\AuthorCitation{Weiwei J.; Jiayun L.}

\address{%
$^{1}$ \quad Department of Electronic Engineering, Tsinghua University, Beijing, China; jiangweiwei@mail.tsinghua.edu.cn\\
$^{2}$ \quad School of Computer Science and Engineering, Nanyang Technological University, Singapore; luoj0028@e.ntu.edu.sg}

\corres{Correspondence: jiangweiwei@mail.tsinghua.edu.cn}

\abstract{Big data have been used widely in many areas including the transportation industry. Using various data sources, traffic states can be well estimated and further predicted to improve the overall operation efficiency. Combined with this trend, this study presents an up-to-date survey of open data and big data tools used for traffic estimation and prediction. Different data types are categorized, and off-the-shelf tools are introduced. To further promote the use of big data for traffic estimation and prediction tasks, challenges and future directions are given for future studies.}

\keyword{Big data; call detail records; census data; GPS trajectory data; location-based service data; open data; public transport transaction data; road sensor data; survey data.}

\begin{document}


\section{Introduction}
\label{sec:introduction}
Big data can be traced back to almost eighty years ago, when people encounter the first attempts to quantify the growth rate in the volume of data or what has popularly been known as the ``information explosion'' (a term first used in 1941, according to the Oxford English Dictionary). The term ``big data'' first appeared in a publication named ``Application-controlled demand paging for out-of-core visualization'' written by Michael Cox and David Ellsworth in October 1997~\citep{cox1997application}. The first definition of big data is ``Visualization provides an interesting challenge for computer systems: data sets are generally quite large, taxing the capacities of main memory, local disk, and even remote disk. We call this the problem of big data.'' With the development of computers, smartphones, the Internet, and sensory equipment, data continue to increase at faster and faster speeds. Broadly speaking, big data can be defined as data sets whose size or type is beyond the ability of traditional relational databases to capture, manage and process with low latency. The characteristics of big data are summarized into five Vs, which represent volume, velocity, variety, veracity, and value~\citep{5v}. The volume represents a large amount of data with unknown value coming from mobile devices, social media, the Internet of Things (IoT) and more. Data can accumulate to tens of terabytes or perhaps hundreds of petabytes. The velocity means the fast rate at which data are received and cumulated and the need for acting on data at an increasing pace. The variety refers to different data types available, including unstructured or semi-structured data such as text, audio and video. These data types often require additional preprocessing. The veracity represents the quality and accuracy of the data. Data in the world can be messy, especially in the case of big data when data dimensions and data sources increase. The last one is value. Data have intrinsic value, but it would have no use if it is not transformed into a usable format or it is unable to extract information from it. With the development of big data, relevant tools are also constantly being developed and updated to fulfill growing needs. Major data processing and analytics tools include Hadoop, Spark, Flink, Cordova, Kafka, and Mahout. Hadoop HDFS, HBase, MongoDB, Hive, and SQL are the in-memory databases used for storing big data. Spark, S4 distributed stream computing platform and Apache Streams are applications that support processing streaming data.

As one of the typical application scenarios, big data has been widely used in the transportation domain. Closely connected to intelligent transportation systems, the emerging infrastructures of the IoT, cyber physical systems (CPS) and smart cities have provided great opportunities for the collection of big data through static sensors, surveillance cameras, and mobile devices. The data size has grown from Trillionbyte to Petabyte levels. Big data have been used for various transportation modes with multiple purposes, e.g., strategic air traffic management~\citep{xie2019similarity}, travel pattern modeling~\citep{gong2016inferring}, road congestion pattern prediction~\citep{he2017mapping}, etc. Big data are used for government policies, e.g., decision-making support tools for smart cities, transparent governance and critical operations. Cutting-edge technologies integrated with big data are also used for urban planning and smart cities, for example, big data, in-memory computing, deep learning, and GPUs (graphics processing units) are used in rapid transit systems~\citep{aqib2019rapid}. For a broader discussion of the applications of big data in the transportation domain, interested readers may refer to the relevant surveys~\citep{torre2018big, moharm2018big, zhu2018big, neilson2019systematic}.

Among different applications, traffic estimation and prediction are the two most significant tasks, which are the focus of this study. While various data related to traffic situations can be obtained, traffic information, e.g., volume, speed, and travel time, is not available without some processing of the raw data in some cases. The process is referred to as the traffic estimation problem, whose target is to extract precise traffic information from any raw data. The aims of traffic estimation include the implementation of a method that can be used to extract precise traffic information and further calibrate a traffic model such as the MMS-model~\citep{allstrom2016highway}~\footnote{MMS-model is a dedicated model name, instead of an abbreviation.}. While estimating the traffic information can only give us the historical states, the aim of traffic prediction is to predict the future situation based on the historical input and adopt appropriate measures accordingly, e.g., traffic control. Various methods have been proposed for traffic estimation and prediction tasks, including statistical models, machine learning models, and deep learning models, in which deep learning models are becoming dominant because they show the best performance~\citep{jiang2018geospatial, jiang2021graph, jiang2021applications, jiang2021graph1}. The success of deep learning models is partially attributed to big data because these models rely on a large training dataset. Several open datasets also boost the development and fair comparison among new models.

To summarize, significant progress has been achieved in previous studies for traffic estimation and prediction with the appearance of big data. The additional benefits of using big traffic data are multi-folds. First, bigger data are the basis of mining longer and more complex patterns hidden in the transportation domain. Second, bigger data make the effective training of artificial intelligence models feasible, especially the deep neural networks, when the data are not due to selection problems and structural changes in data are not considered. Last, bigger data from various sources are the basis of capturing the relationship among different transportation systems. However, there is a lack of an up-to-date summary and collection of open datasets and tools. Some of the relevant studies are based on private data, whose results are impossible to replicate. In this survey, we focus on open datasets, especially large-volume and multi-modal datasets. To further boost the relevant studies, we also release a processed GPS trajectory dataset that is collected from more than 20,000 taxi drivers in Beijing in three months, namely, November 2012, November 2014 and November 2015, which has been used in our previous studies~\citep{jiang2017multi, jiang2018impact}. The dataset is publicly available~\footnote{The data would be available here: \url{https://github.com/jwwthu/DL4Traffic}}.

Our contributions in this paper are summarized as follows:
\begin{itemize}
\item We summarize the different data types used for traffic estimation and prediction tasks;
\item We summarize the latest collection of relevant open datasets;
\item We contribute a new GPS trajectory dataset for the research community;
\item We summarize the collection of relevant big data tools;
\item We point out the challenges and future directions of utilizing big data for traffic estimation and prediction tasks.
\end{itemize}

The following parts of this paper are organized as follows. The data used for traffic estimation and prediction tasks are summarized in Section~\ref{sec:data}. The big data tools are collected and introduced in Section~\ref{sec:tools}. The relevant challenges and future directions are pointed out in Section~\ref{sec:challenge}. The conclusion is drawn in Section~\ref{sec:conclusion}.

\section{Data-driven Traffic Estimation and Prediction}
\label{sec:data}

In this section, we summarize the different types of data that can be used for traffic estimation and prediction. We also contribute an up-to-date collection of available open datasets for each data type as well as new GPS trajectory data for further studies. There are different ways of categorizing traffic big data. For example, traffic big data were previously divided into supervised and unsupervised types, in which supervised data are direct sources of traffic information, e.g., loop detectors and GPS traces, while unsupervised data are indirect sources that can be used to infer traffic information, e.g., call detail records and cell-phone position data. In this study, we further divide traffic big data into more types based on data sources.

Open data policies in different countries vary greatly. Take the U.S. government as an example, transportation related open data can be found from the National Transit Database (NTD)~\footnote{\url{https://www.transit.dot.gov/ntd/ntd-data}}, Federal Highway Administration (FHWA)~\footnote{For example, Urban Congestion Reports~\url{https://ops.fhwa.dot.gov/perf_measurement/ucr/}}, Bureau of Economic Analysis (BEA)~\footnote{\url{https://www.bea.gov/data/}}, and American Community Survey (ACS)~\footnote{\url{https://www.census.gov/programs-surveys/acs/data.html}}. The Next Generation Simulation (NGSIM)~\footnote{\url{https://ops.fhwa.dot.gov/trafficanalysistools/ngsim.htm}} is the most widely used open-source vehicle trajectory dataset for traffic flow studies~\citep{li2020trajectory}. In this study, we mainly focus on open datasets in academia. More data collections can be found online, which may be maintained by individuals and research institutes, e.g., mobility datasets~\footnote{\url{https://privamov.github.io/accio/docs/datasets.html}}, open traffic collection~\footnote{\url{https://github.com/graphhopper/open-traffic-collection/}}, and Beijing City Lab~\footnote{\url{https://www.beijingcitylab.com/data-released-1/}}.

\subsection{Trip surveys}
Trip surveys are detailed questionnaires on mobility habits, which are usually collected by local authorities or researchers. Sometimes, it is only a way to accurately measure and understand people's changing daily travel patterns, when different travel modes are considered and location privacy is not violated. For example, in the "My Daily Travel Survey"~\footnote{\url{https://www.cmap.illinois.gov/data/transportation/travel-survey\#My_Daily_Travel_Survey}} by the Chicago Metropolitan Agency for Planning between August 2018 and April 2019, households in northeastern Illinois are asked to tell the trips they made for work, school, shopping, errands, and socializing with family and friends. 

Another example is the "California Household Travel Survey (CHTS)"~\footnote{\url{https://www.nrel.gov/transportation/secure-transportation-data/tsdc-california-travel-survey.html}} by the National Renewable Energy Laboratory (NREL) between 2010 and 2012. As the largest such regional or statewise survey ever conducted in the United States, detailed travel behavior information was obtained from more than 42,500 households via multiple data-collection methods, including computer-assisted telephone interviewing, online and mail surveys, wearable (7,574 participants) and in-vehicle (2,910 vehicles) global positioning system (GPS) devices, and on-board diagnostic sensors that gathered data directly from a vehicle's engine. Details of personal travel behavior were gathered within the region of residence, inter-regionally within the state, and in adjoining states and Mexico. The survey sampling plan was designed to ensure an accurate representation of the entire population of the state. Among the participating households, 42,436 completed the travel diary survey portion, 3,871 completed the wearable GPS portion, and 1,866 completed the vehicle GPS portion of the study. Trip details including purpose, mode, travelers, tolls, departure time, arrival time, and distance are collected for both private vehicles and public transit trips. More trip surveys are available on the Internet~\footnote{\url{https://www.nrel.gov/transportation/secure-transportation-data/tsdc-cleansed-data.html}}.

The pros of trip surveys include high-resolution and detailed information, which contain trip information directly and eliminates the need for traffic estimation. The cons of trip surveys include small sample size, limited spatial and temporal scale, self-reporting errors, and high cost to collect, which limits the availability of such data as well as real-time applications, e.g., they are less used for traffic prediction.

\subsection{Census Data and Survey Data}
Census data and survey data may also contain traffic information and other information useful for traffic problems, such as locations of residence and workplace. These data are usually collected by governments periodically. For example, in the Commuting Flows~\footnote{\url{https://www.census.gov/topics/employment/commuting/guidance/flows.html}} by the US Census Bureau, the primary workplace location from the respondents are collected. In the Migration Data~\footnote{\url{https://www.irs.gov/statistics/soi-tax-stats-migration-data}} by the Internal Revenue Service, both inflows and outflows of migration patterns are available for Filing Years 1991 through 2018 with the IRS. More survey data can be found in Table~\ref{tab:data_statistics}.

\begin{table}[H]
\caption{The list of open survey data.\label{tab:data_statistics}}
    \begin{adjustwidth}{-\extralength}{0cm}
        \begin{tabularx}{\fulllength}{|p{3.3cm}|p{1.8cm}|p{2cm}|p{2.7cm}|p{6.5cm}|}
            \toprule
            Name & Type & Spatial Range & Temporal Range & Download Link \\
            \midrule
            GB Road Traffic Counts & Aggregated traffic counts & Great Britain & 2000-2019 & \url{https://data.gov.uk/dataset/208c0e7b-353f-4e2d-8b7a-1a7118467acc/gb-road-traffic-counts} \\
            \hline
            Highways England network journey time and traffic flow data & Traffic flow, travel time & Great Britain & 2006-2020 & \url{https://data.gov.uk/dataset/9562c512-4a0b-45ee-b6ad-afc0f99b841f/highways-england-network-journey-time-and-traffic-flow-data} \\
            \hline
            DataMall & Multiple types & Singapore & Frequently updated & \url{https://datamall.lta.gov.sg/content/datamall/en/dynamic-data.html} \\
            \hline
            Minnesota Department of Transportation Traffic Data & Traffic flow, traffic speed & Twin Cities, Minnesota, USA & Since 2011 & \url{http://data.dot.state.mn.us/datatools/} \\
            \hline
            Chicago Traffic Tracker & Traffic speed & Chicago, USA & 2011-2018 & \url{https://data.cityofchicago.org/Transportation/Chicago-Traffic-Tracker-Historical-Congestion-Esti/77hq-huss/data} \\
            \hline
            US-Accidents~\citep{moosavi2019countrywide} & Traffic accident & USA & Feb. 2016 to Dec. 2020 & \url{https://smoosavi.org/datasets/us_accidents} \\
            \bottomrule
        \end{tabularx}
    \end{adjustwidth}
\end{table}

The pros of census data and survey data are their large sample size and large coverage, which is usually the entire country. The cons are also obvious. The flows are aggregated at the municipality or county level, which is coarse. There is only limited traffic information. It is also expensive and time consuming to collect this type of data as trip surveys. Census data and survey data are not widely used for traffic estimation and prediction tasks, with only a few exceptions~\citep{katranji2020deep}.

\subsection{Road Sensor Data}
Static sensors can be deployed to collect traffic data, e.g., inductive loop detectors, magnetic and pneumatic tube sensors, radar sensors, infrared sensors, and acoustic sensors. Loop detectors are the most mature technology among road sensors and contribute many famous open datasets for traffic estimation and traffic prediction. A complete list of open road sensor data used in previous studies is shown in Table~\ref{tab:data_sensor}. The pros of road sensor data are their abilities to collect large volume data continuously and automatically, usually in minutes or seconds. The cons of these data are their fixed coverage in the spatial range as well as the missing data problem caused by sensor failures. Additionally, the traffic speed obtained from traffic sensors could be a point measurement and may not be suitable to calculate the average travel time across the road link.

\begin{table}[H]
\caption{The list of open road sensor data.\label{tab:data_sensor}}
    \begin{adjustwidth}{-\extralength}{0cm}
        \begin{tabularx}{\fulllength}{|p{4.2cm}|p{1.8cm}|p{2.6cm}|p{2.8cm}|p{5cm}|}
            \toprule
            Name & Type & Spatial Range & Temporal Range & Download Link \\
            \midrule
            Performance Measurement System (PeMS) Data & Traffic speed, traffic flow & California, USA & 2001-2019 & \url{http://pems.dot.ca.gov/} \\
\hline
METR-LA~\citep{li2018diffusion} & Traffic speed, traffic flow & Los Angeles, USA & March 1st to June 30th, 2012 & \url{https://github.com/liyaguang/DCRNN} \\
\hline
Seattle-Loop-Data~\citep{cui2018deep} & Traffic Speed & Seattle, USA & January 1st to 31st, 2015 & \url{https://github.com/zhiyongc/Seattle-Loop-Data} \\
\hline
Traffic Speed Guangzhou~\citep{xinyu_chen_2018_1205229} & Traffic speed & Guangzhou, China & August 1, 2016 to September 30, 2016 & \url{https://zenodo.org/record/1205229} \\
\hline
Traffic Flow Madrid & Traffic flow & Madrid, Spain & Since 2013 & \url{http://datos.madrid.es} \\
\hline
Traffic Flow Whitemud Drive & Traffic flow & Whitemud Drive, Canada & August 6, 2015 to August 28, 2015 & \url{http://www.openits.cn/openData1/700.jhtml} \\
\hline
UTD19~\citep{loder2019understanding} & Traffic capacity & 40 cities globally & Various time periods & \url{https://utd19.ethz.ch/index.html} \\
\hline
PORTAL & Traffic speed & Portland-Vancouver Metropolitan region, USA & Regularly updated & \url{https://portal.its.pdx.edu/home} \\
            \bottomrule
        \end{tabularx}
    \end{adjustwidth}
\end{table}

Traffic signals also belong to road sensor data, which are often seen in the literature. Different types of traffic signal data can be collected in traffic light, traffic signal lantern, traffic signal controller box, traffic signal pole, etc. For example, traffic light data are collected in those traffic lights with an Internet connection, with an typical data type of numeric values for time to green. They also contain location and attributes of traffic controls located at each intersection. This type of data is more often used for traffic control relevant studies, instead of traffic estimation or prediction. For example, an open collection of the traffic light data is provided in~\citep{signal} for the study of reinforcement learning based traffic control.

\subsection{Call Detail Records}
Call detail records (CDRs), cellular signaling data and cell phone position data are all generated from the cellular network, which contains similar information on human mobility patterns. These data are produced by a phone carrier to store details of calls passing through a device and usually contain various attributes of the call, e.g., timestamp, source, destination, base transceiver station (BTS), duration. Then the position and trajectory (sequence of locations) can be inferred from CDRs. Compared with GPS trajectory data with a higher spatial resolution, CDRs provide a coarse positioning approach. Data preprocessing is necessary not only for removing noise but also for identifying the key locations. As one of the pioneering studies, cellular phone tracking data are used for traffic volume estimation and further evaluated with loop detector data in~\cite{caceres2012traffic}. Afterwards, an increasing number of studies have used call detail records and cell phone position data for traffic status inference, even in recent years. One-month cellular signaling data (SD) are used to extract road-level human mobility with a multi-information fusion framework, which takes the SD uncertainty issue into consideration~\citep{song2020miff}. Based on a regression model, a cell probe (CP)-based method is proposed to estimate the vehicle speed with the normal location update (NLU) procedure and the consecutive handoff (HO) event as inputs, and the proposed method achieves a 97.63\% accuracy~\citep{chen2020cell}. Long-term evolution (LTE) access data are used as input for a deep-learning-based road traffic prediction system, to predict the real-time speed of traffic~\citep{ji2019deep}. Characteristics of human mobility patterns are revealed by high-frequency cell phone position data in~\cite{zhao2021characteristics}. More usages can be found in recent surveys~\citep{ghahramani2020urban}.

The publicly available CDR or similar data are shown in Table~\ref{tab:data_cdr}. The pros of using CDR or similar data are their ubiquitous availability (with mobile phones), large volume, and multiple dimensions (social, mobile, time, demographics). However, the cons include poor positioning ability, noise, and preprocessing requirements. Because the position can only be derived from the cellular BTSs, it is partially detected only when calls are made and known at the BTS level only. Moreover, the ping-pong effect between different BTSs can create noise in the collected data. Another concern when using CDR data is the potential privacy leakage issue for both the personal and location information. Several privacy protection solutions have been proposed. The personal information is anonymized and encrypted, with a minimal possibility of tracing back to the individuals. The location information can be aggregated in a larger region, without revealing the precise coordinates.

\begin{table}[H]
\caption{This is a wide table.\label{tab:data_cdr}}
    \begin{adjustwidth}{-\extralength}{0cm}
        \begin{tabularx}{\fulllength}{|p{3.8cm}|p{3.5cm}|p{3cm}|p{6.5cm}|}
            \toprule
            Name & Spatial Range & Temporal Range & Download Link \\
            \midrule
Telecommunications-SMS, call, internet-MI~\citep{italia2015telecommunications} & Milan, Italy & November 1, 2013 to Januray 1, 2014 & \url{https://dataverse.harvard.edu/dataset.xhtml?persistentId=doi:10.7910/DVN/EGZHFV} \\
\hline
OpenCellID & Multiple cities globally & Regularly updated & \url{https://www.opencellid.org} \\
            \bottomrule
        \end{tabularx}
    \end{adjustwidth}
\end{table}

\subsection{GPS Trajectory Data}
GPS trajectory data are often collected from floating vehicles, which are equipped with on-board positioning systems and communication devices. Additionally, smartphones can also be used for GPS trajectory data collection, e.g., in a crowd-sourced approach. While GPS trajectory data are easy to collect, the preporcessing steps are necessary, e.g., location detection to group points into one meaningful location and trajectory segmentation to split a trajectory in sub-trajectories. Map matching is also used to map the GPS coordinates into the road network. Because of advantages such as fine granularity, GPS-based “Track and Trace” data have been formally defined and highlighted for transportation modeling and policy-making in a recent survey~\citep{harrison2020new}. Taxi trajectory data can also be used for trip purpose inference and travel pattern discovery~\citep{gong2016inferring}. More relevant studies are using GPS trajectory data for traffic estimation and prediction, e.g., travel time is estimated with the GPS trajectory data in~\cite{zou2020estimation} by combining the gradient boosting decision tree (GBDT) model and the deep neural network (DNN) model, and OD pairs are predicted with big GPS data in~\cite{wang2020urban}.

Due to privacy concerns, most of the open GPS trajectory data are collected from taxis, with only a few exceptions, as shown in Table~\ref{tab:data_gps}. Additionally, in many cases, the raw GPS trajectory data are not shared, and only the estimated traffic states derived from the GPS trajectory are publicly available. The pros of using GPS trajectory data include the high spatial and temporal resolution as well as the tracking ability of the full traces. Taxis can also be seen as mobile probes, which reflect all road traffic situations. The cons are similar to CDRs, which include the preprocessing requirement and the noise in the data. Moreover, the GPS trajectory data are often collected at a high frequency, e.g., in seconds, so the huge volume of such data requires storage and computation abilities, which can only be fulfilled by big data tools.

\begin{table}[H]
\caption{The list of open GPS trajectory and relevant data.\label{tab:data_gps}}
    \begin{adjustwidth}{-\extralength}{0cm}
        \begin{tabularx}{\fulllength}{|p{3.4cm}|p{2.7cm}|p{1.8cm}|p{2.4cm}|p{6cm}|}
            \toprule
            Name & Type & Spatial Range & Temporal Range & Download Link \\
            \midrule
SF Taxis or Cabspotting~\citep{piorkowski2009crawdad} & Taxi GPS trajectory & San Francisco, USA & May 17, 2008 to June 10, 2008 & \url{http://crawdad.org/epfl/mobility/20090224/index.html} \\
\hline
Rome Taxis~\citep{bracciale2014crawdad} & Taxi GPS trajectory & Rome, Italy & Feburary 1, 2014 to March 3, 2014 & \url{http://crawdad.org/roma/taxi/20140717/index.html} \\
\hline
Porto Taxis & Taxi GPS trajectory & Porto, Portugal & July 1, 2013 to June 30, 2014 & \url{https://www.kaggle.com/c/pkdd-15-predict-taxi-service-trajectory-i/data} \\
\hline
Geolife~\citep{zheng2010geolife} & Taxi GPS trajectory & Beijing, China & Apr. 2007 to Aug. 2012 & \url{https://www.microsoft.com/en-us/download/details.aspx?id=52367} \\
\hline
Mobile Data Challenge (MDC)~\citep{kiukkonen2010towards} & Taxi GPS trajectory & Lake Geneva region & 2009 to 2011 & \url{https://www.idiap.ch/dataset/mdc} \\
\hline
TaxiCD & Taxi GPS trajectory & Chengdu, China & August 3rd to 30th, 2014 & \url{https://js.dclab.run/v2/cmptDetail.html?id=175} \\
\hline
Grab-Posisi~\citep{huang2019grab} & Grab Drive GPS trajectory & Southest Asia & April 8, 2019 to April 21, 2019 & Upon request \\
\hline
PrivateCarTrajectoryData & Private car GPS trajectory & Shenzhen, China & January 2016 & \url{https://github.com/HunanUniversityZhuXiao/PrivateCarTrajectoryData} \\
\hline
TaxiBJ~\citep{zhang2017deep} & Taxi traffic flow & Beijing, China & Multiple time periods & \url{https://github.com/Mouradost/ASTIR} or \url{https://www.jianguoyun.com/p/DesHv2UQs-HRBxi5gtYB} \\
\hline
BikeNYC~\citep{mourad2019astir} & Bike traffic flow & NYC, USA & April 1, 2014 to Sepetmber 30, 2014 & \url{https://github.com/Mouradost/ASTIR} or \url{https://www.jianguoyun.com/p/DesHv2UQs-HRBxi5gtYB} \\
\hline
Traffic Speed Chengdu~\citep{guo2019urban} & Taxi traffic speed & Chengdu, China & June 1, 2015 to July 15, 2015 & \url{https://doi.org/10.6084/m9.figshare.7140209.v4} \\
\hline
SHSpeed (Shanghai Traffic Speed)~\citep{wang2018efficient} & Taxi traffic speed & Shanghai, China & April 1st to 30th, 2015 & \url{https://github.com/xxArbiter/grnn} \\
\hline
TaxiSZ~\citep{zhao2019t} & Taxi traffic speed & Shenzhen, China & January 1st to 31st, 2015 & \url{https://github.com/lehaifeng/T-GCN} \\
            \bottomrule
        \end{tabularx}
    \end{adjustwidth}
\end{table}

While taxi GPS trajectory datasets are widely used in the literature, there is a concern that taxi drivers are not the best representatives since they are experienced drivers with a higher possibility of finding optimal routes. In some studies, this wisdom from taxi drivers are used to built better navigation services~\citep{zheng2010drive}. On the contrary, the driving behavior gap between taxi drivers and private car drivers are becoming smaller when these navigation services are now widely used by both groups. In the studies for traffic estimation and prediction, taxi drivers are preferred as probes because they drive longer and provide more traffic situation measurements with a larger coverage.

In addition to the existing datasets, we also contribute a new GPS trajectory dataset for the research community in this study. The GPS trajectory data are collected in Beijing during three time periods, namely, November 2012, November 2014, and November 2015. Each GPS data sample contains the following attributes: anonymous taxi identity, timestamp, latitude, longitude, azimuth, spot speed, and operation status (occupied, vacant or stopped). The data are sampled with an interval of approximately one minute. The data summary is shown in Table~\ref{tab:our_gps}. This dataset is publicly available~\footnote{The data would be available here: \url{https://github.com/jwwthu/DL4Traffic}}.

\begin{table}[!htb]
    \centering
    \caption{Data summary for the GPS trajectory dataset contributed in this study.}
    \begin{tabular}{|l|c|c|c|}
        \hline
        Time Periods & November 2012 & November 2014 & November 2015  \\
        \hline
        Taxi Drivers & 8,879 & 17,749 & 20,067 \\
        \hline
        Days & 30 & 30 & 30 \\
        \hline
    \end{tabular}
    \label{tab:our_gps}
\end{table}

\subsection{Location-Based Service Data}
With the development of the mobile Internet, location-based service (LBS) data have arisen with GPS functionality on smartphones. Various location-based big data are collected from location-based social networks, e.g, checkins, geotagged tweets and micro-blogs, and Maps and Navigation Apps, e.g., Google Map and Baidu Map. These data are often collected in a crowd-sourced approach, and a traffic information extraction step is necessary. To effectively collect, integrate and process crowd-sourced data including mobile applications, webs, and external data sources, a prototype system is developed and validated in~\cite{mai2020mining}, to infer traffic conditions. Traffic speed data of 29 cities across the world over a 40-day period are gathered from Google Map API and further used for the analysis of traffic congestion patterns~\citep{nair2019characterizing}. Social media texts may also contain traffic information, e.g., geotagged tweets and micro-blogs. However, the challenge is to extract traffic-relevant information from natural languages. Deep learning models have been proposed for conducting information extraction, e.g., an LSTM-CNN is proposed to extract traffic-relevant microblogs in~\cite{chen2018detecting}, which outperforms the baselines including the support vector machine model based on a bag of n-gram features. Social media data are further proven effective for traffic accident detection and reporting~\citep{wan2020empowering, ali2021traffic}, traffic jam management~\citep{wang2020traffic}. Combining social media data from multiple social networks can further improve the traffic event detection accuracy, e.g., the case of combining Arabic and English data streams from Twitter and Instagram used in the SNSJam system~\citep{alkouz2020snsjam}. However, location-based big data may be low-quality with noise. Taking Bluetooth speed data as the ground truth, the quality of Waze data in Sevierville, TN, USA is evaluated in~\cite{hoseinzadeh2020quality}, in which a kNN method manages to achieve a prediction accuracy of 84.5\% and 82.9\% for traffic speed estimation based on Waze data as the input.

The list of open location-based service data is shown in Table~\ref{tab:data_lbs}. The pros of using LBS data include their wide availability and semantic information, e.g., restaurants and malls. However, the cons are also obvious. Human mobility can only be partially detected when check-ins or geotagged texts are made. The data are thus very sparse and can be sparser than CDRs. There is also self-selection bias, which would cause inaccurate traffic information.

\begin{table}[H]
\caption{The list of open location-based service data.\label{tab:data_lbs}}
    \begin{adjustwidth}{-\extralength}{0cm}
        \begin{tabularx}{\fulllength}{|p{3cm}|p{2.9cm}|p{2.2cm}|p{2.6cm}|p{5.5cm}|}
            \toprule
            Name & Type & Spatial Range & Temporal Range & Download Link \\
            \midrule
Q-Traffic or BaiduTraffic~\citep{liao2018deep} & Traffic speed from navigation apps & Beijing, China & April 1, 2017 to May 31, 2017 & \url{https://github.com/JingqingZ/BaiduTraffic} \\
\hline
Alibaba Cloud Tianchi Dataset & Travel time from navigation Apps & Guiyang, China & Apr. 2017 & \url{https://tianchi.aliyun.com/competition/entrance/231598/information} \\
\hline
Brightkite~\citep{cho2011friendship} & Check-ins & N/A & Apr. 2008 to Oct. 2010 & \url{https://snap.stanford.edu/data/loc-brightkite.html} \\
\hline
Gowalla~\citep{cho2011friendship} & Check-ins & N/A & Feb. 2009 to Oct. 2010 & \url{https://snap.stanford.edu/data/loc-gowalla.html} \\
\hline
Foursquare~\citep{yang2013sentiment} & Check-ins & NYC, USA & October 24, 2011 to February 20, 2012 & \url{https://sites.google.com/site/yangdingqi/home/foursquare-dataset} \\
\hline
Yelp & Check-ins & Multiple cities globally & Regularly updated & \url{https://www.yelp.com/dataset}\\
\hline
MapBJ~\citep{cheng2018deeptransport} & Traffic congestion from navigation Apps & Beijing, China & Mar. 2016 to June 2016 & \url{https://github.com/cxysteven/MapBJ} \\
\hline
Tecent API & Traffic flow index & China & Since 2015 & \url{https://heat.qq.com/} \\
\hline
Uber Movement & Travel time and speed & Multiple cities globally & Since 2017 & \url{https://movement.uber.com/} \\
            \bottomrule
        \end{tabularx}
    \end{adjustwidth}
\end{table}

\subsection{Public Transport Transaction Data}
Transaction data can be collected in various public transport systems, especially those with automatic fare collection (AFC) systems, and further used for traffic estimation and prediction. For example, smart card data are often used for public transit origin-destination (OD) estimation~\citep{hussain2021transit}. Based on smart card and bus trajectory data, a two-stage transportation analysis approach is proposed to reconstruct the individual passenger trips and cluster these trips to identify the transit corridors in~\cite{zhang2020identifying}. Based on 10 million taxi trip records in New York and Shenzhen and using spatio-temporal clustering, Bayesian probability and Monte Carlo simulation, a two-layer framework, which consists of an activity inference model and a pairing journey model, is proposed to extract and predict travel patterns~\citep{gong2020extracting}. Using the gravity model and Bayesian rules, the purpose of dockless shared-bike users is inferred from a shared bike dataset in Shenzhen, China, and the introduction of activity type proportion and service capacity of point of interest (POIs) is proven effective for inference~\citep{li2021inferring}. Based on daily OD amount by transportation and by purpose and several surveys on time use and activities, an open people mass movement dataset named Open PFLOW is built in~\cite{kashiyama2017open}, which achieves a comparable accuracy with other approaches, e.g., commercial mobility data and traffic census.

Another approach of collecting public transport transaction data is based on Internet of Things and counter devices installed in the vehicle doors, e.g., infrared or RGB image-based passenger counters. This passive approach of collecting data can be conducted silently, without the burden of passenger involvement. However, more errors may exist in the collected due to the device fault or misjudgement caused by passenger's strange behaviors. 

There are also many public transport transaction data that are available for the research community as shown in Table~\ref{tab:data_transaction}. The pros of using public transport transaction data are their wide availability and close connection with traffic states. The cons are the storage and computation requirements, which can be high if the data cover wide spatial and temporal ranges.

Another problem with public transport transaction data is that they can be incomplete. For example, in some countries and regions, tickets are only validated at the entrance to the vehicles, e.g., bus or subway. In those cases, only the inflow data can be collected and the scope of follow-up studies is limited, without being able to obtain the outflow or OD flow situations.

\begin{table}[H]
\caption{The list of open public transport transaction data.\label{tab:data_transaction}}
    \begin{adjustwidth}{-\extralength}{0cm}
        \begin{tabularx}{\fulllength}{|p{2.5cm}|p{2cm}|p{2.5cm}|p{3cm}|p{6.2cm}|}
            \toprule
            Name & Type & Spatial Range & Temporal Range & Download Link \\
            \midrule
SHMetro~\citep{liu2020physical} & Metro & Shanghai, China & July 1, 2016 to September 30, 2016 & \url{https://github.com/ivechan/PVCGN} \\
\hline
HZMetro~\citep{liu2020physical} & Metro & Hangzhou, China & January 2019 & \url{https://github.com/ivechan/PVCGN} \\
\hline
Hangzhou Metro & Metro & Hangzhou, China & January 1, 2019 to January 25, 2019 & \url{https://tianchi.aliyun.com/competition/entrance/231708/information} \\
\hline
Bike Bay Area~\citep{yi2019rebalancing} & Shared bike & Bay Area, USA & September 1, 2015 to August 31, 2016 & \url{https://github.com/TwinkleBill/babs_open_data_year_3} \\
\hline
BikeNYC & Shared bike & NYC, USA & July 1, 2013 to December 12, 2016 & \url{https://www.citibikenyc.com/system-data} \\
\hline
BikeDC & Shared bike & Washington D.C., USA & 2011-2016 & \url{https://www.capitalbikeshare.com/system-data} \\
\hline
BikeChicago & Shared bike & Chicago, USA & 2015-2020 & \url{https://www.divvybikes.com/system-data} \\
\hline
Bike Chattanooga Trip Data & Shared bike & Chattanooga, Tennessee, USA & July 23, 2012 to April 9, 2020 & \url{https://data.chattlibrary.org/} \\
\hline
Mobike Beijing~\citep{cao2020analysis} & Shared bike & Beijing, China & May 10, 2017 to May 24, 2017 & \url{https://github.com/SharingBikeNNU/Riding-Modes_Tucker} \\
\hline
Ride Austin & Ride sharing & Austin, USA & June 2, 2016 to April 13, 2017 & \url{https://data.world/ride-austin} \\
\hline
UberNYC & Ride sharing & NYC, USA & from April to September 2014 & \url{https://github.com/fivethirtyeight/uber-tlc-foil-response} \\
\hline
Didi GAIA Open Data & Ride sharing & Chengdu, Xi'an, and Haikou, China & Various time periods & \url{https://outreach.didichuxing.com/research/opendata/} \\
\hline
TaxiNYC & Taxi & NYC, USA & Since 2009 & \url{http://www.nyc.gov/html/tlc/html/about/trip_record_data.shtml} \\
            \bottomrule
        \end{tabularx}
    \end{adjustwidth}
\end{table}

\subsection{Surveillance and Airborne Digital Cameras}
Closed circuit television (CCTV) cameras are widely used for monitoring traffic patterns and helping police forces in large cities. By collecting these surveillance videos, traffic information can be estimated and further used for prediction. For example, traffic states with CCTV systems are analyzed in~\cite{bui2020multi}. However, the challenge is the heavy workload of processing digital multimedia data. Transportation video data management is considered in~\cite{hao2020design}, and a high-performance computing architecture is developed with distributed files and distributed computing systems. Edge computing is also applied for traffic flow detection in real-time videos. A specific edge device Jetson TX2 platform is used in~\cite{chen2020edge}, with the YOLOv3 (You Only Look Once) model for vehicle detection and the optimized DeepSORT (Deep Simple Online and Realtime Tracking) algorithm for vehicle tracking. The edge device manages to achieve an average processing speed of 37.9 frames per second, with an average accuracy of 92.0\%. In addition to static traffic surveillance cameras, remote sensing tools, e.g., airborne digital cameras and unmanned aerial vehicle (UAV) cameras, can also be used for traffic monitoring. For example, the average traffic speed and density are estimated with an airborne optical 3K camera system in~\cite{leitloff2014operational}, and the traffic flow parameter is calculated from UAV aerial videos in~\cite{brkic2020analytical}.

Some open video data are summarized in Table~\ref{tab:data_video}. The pros of video data are their continuous observation, wide coverage for road traffic, and multiple usages for traffic estimation, prediction, accident analysis and even vehicle tracking. The cons of video data are their high storage and computation requirements. The video stream also has many engineering challenges, e.g., compression artifacts, blurring, and hardware faults.

\begin{table}[H]
\caption{The list of open video data.\label{tab:data_video}}
    \begin{adjustwidth}{-\extralength}{0cm}
        \begin{tabularx}{\fulllength}{|p{4.5cm}|p{3.2cm}|p{3cm}|p{6cm}|}
            \toprule
            Name & Spatial Range & Temporal Range & Download Link \\
            \midrule
MIT Traffic Dataset~\citep{wang2008unsupervised} & One camera in MIT & 90 minutes long & \url{http://mmlab.ie.cuhk.edu.hk/datasets/mit_traffic/index.html} \\
\hline
Video Surveillance Data~\citep{fedorov2019traffic} & One camera in Chelyabinsk, Russia & 982 video frames & \url{https://github.com/alnfedorov/traffic-analysis} \\
            \bottomrule
        \end{tabularx}
    \end{adjustwidth}
\end{table}

\subsection{License-plate Recognition Data}
License plate recognition (LPR) data are emerging data sources for vehicle re-identification as well as traffic estimation and prediction. Large-scale LPR data collection systems are being developed all over the world, e.g., the automatic number-plate recognition (ANPR) systems. However, due to data privacy concerns, there are few open datasets and few relevant studies. In~\cite{zhan2020link}, a six-day LPR dataset from a small road network in Langfang, China, is used to estimate and predict link-based traffic states, and the feasibility is validated by using a more comprehensive link-level field experiment dataset. A similar type of data is electronic registration identification (ERI), which is also used in traffic flow prediction~\citep{zheng2020dynamic}. 

The pros of LPR or ERI data are their precise tracking abilities of individual vehicles. However, this feature also causes data privacy concerns. Another disadvantage is that there are no publicly available LPR or ERI datasets.

\subsection{Toll Ticket Data}
Electronic toll collection (ETC) systems are widely applied in toll roads, HOV lanes, toll bridges, etc. For example, more than 90 percent of vehicles on expressways in China were equipped with the ETC system by the end of 2019. Toll ticket data (TTD) collected from ETC systems provide a high-coverage and low-cost approach for expressway traffic estimation. Based on the TTD from the Shandong Expressway Toll System in China, a simulation-based dynamic traffic assignment algorithm is proposed to obtain traffic flow in~\cite{yao2021traffic} and achieves a high examination accuracy (approximately 5\% MAPE). The only public source of toll ticket data used for traffic flow prediction as the authors know is the Amap data for KDD CUP 2017~\footnote{\url{https://tianchi.aliyun.com/dataset/dataDetail?dataId=60}}.

The pros of tool ticket data include their high coverage and low cost. However, the cons are obvious. Toll ticket data can only be collected in places where ETC systems are applicable.

\subsection{External Data}
In addition to the above data that are used to extract and predict traffic states, some external data are also used for these problems as supplementary data, e.g., weather data, calendar information, emissions of air pollutants, and noise. These data do not have a large volume and are easily available. There are many public sources for weather data, e.g., Weather Underground API~\footnote{\url{https://www.wunderground.com/weather/api/}}, MesoWest~\footnote{\url{http://mesowest.utah.edu/}}. Calendar information is also widely accessible from the Internet, e.g., OfficeHolidays~\footnote{\url{https://www.officeholidays.com/}}. To name some relevant studies that use these external data, 48 weather forecasting factors are analyzed and used for traffic flow prediction based on a regression model in~\cite{lee2015prediction}, and rainfall data are also used for similar purposes~\citep{dunne2013weather, jia2017traffic}.

This section is concluded with Table~\ref{tab:data_summary}, which shows the relationship between the potential data sources with possible data attributes. This table is far from being perfect but as a preliminary reference.

\startlandscape
\begin{table}[!htb]
    \centering
    \caption{A summary for the linking between data sources and possible results.}
    \begin{tabular}{|p{3.8cm}|p{2.4cm}|p{2.4cm}|p{1.5cm}|p{1.5cm}|p{3.8cm}|p{1.5cm}|p{1.5cm}|p{1.5cm}|}
        \hline
         & Road Traffic Volume & Road Traffic Speed & Cyclist Speed & Bicycle Volume & Passenger Flow in Public Transport Systems & Pedestrian Speed & Pedestrian Volume & Pedestrian Route \\
        \hline
        Traffic signal controllers \& Road sensors & + & + & & & & & & \\
        \hline
        GPS sensors & + & + & & & & & & \\
        \hline
        Bicycle counters & & & & + & & & & \\
        \hline
        Passenger counters \& AFC systems & & & & & + & & & \\
        \hline
        On-board computers & & & & & + & & & \\
        \hline
        Cellular phones \& Smartphones & & & & & & + & + & + \\
        \hline
        CCTV cameras & + & + & + & + & + & + & + & + \\
        \hline
        ANPR systems & + & +1) & & & & & & \\
        \hline
        ETC systems & +2) & & & & & & & \\
        \hline
        \multicolumn{9}{l}{1) only segmental speed measurement} \\
        \multicolumn{9}{l}{2) only volume measurement in the entrance} \\
    \end{tabular}
    \label{tab:data_summary}
\end{table}
\finishlandscape

\section{Big Data Tools}
\label{sec:tools}
There are already some preliminary trials of applying the latest big data tools for traffic big data, as traditional big data tools are not optimized for big data in the transportation domain. Some preliminary efforts have been proposed for expanding the off-the-shelf tools for a better support of spatiotemporal big data, e.g., Hadoop is expanded with the capacities of spatiotemporal indexing~\citep{li2017high} and trajectory analytics~\citep{bakli2019hadooptrajectory}. Based on PostgreSQL and PostGIS, an open-source mobility database named MobilityDB~\citep{MobilityDB} is proposed for moving object geospatial trajectories, e.g., GPS traces~\citep{zimanyi2020mobilitydb}. For analyzing and visualizing spatiotemporal big data, new data processing algorithms and methods should be implemented as new or extensions of existing commercial or open-source platforms, for ease of use and a better integration, e.g., the latest SuperMap GIS platform provides full support to the Spark computing framework~\citep{wang2019integrated}. Offline and online trajectory analyses are often separated in existing systems. To overcome this shortcoming, a Spark-based hybrid and efficient framework named Dragoon~\citep{fang2021dragoon} is proposed for both offline and online trajectory data analytics. Dragoon manages to decrease storage overhead up to doubled compared with the state-the-art big trajectory data management system Ultraman~\citep{ding2018ultraman}, while maintaining a similar offline trajectory query performance. Dragoon also manages to achieve at least 40\% improvement of scalability over Flink and Spark Streaming and achieves an average doubled performance improvement for online trajectory data analytics. Based on cellular data (2G/3G/4G), base station data, user information and road network data, a real-time urban mobility monitoring and traffic management system is proposed in~\cite{yao2020understanding}, leveraging big data tools including Hive, Spark, and Hbase. The proposed system proposes a total of 600 TB cellphone data collected over 3 million people daily for 3 years in a field case study in Guiyang, China.

In this section, we collect and summarize the off-the-shelf big data storage and computing tools that are already used or can be further leveraged in traffic estimation and prediction tasks. While these tools have not been widely used yet, there is already the trend of more and more data being collected and used in traffic problems, e.g., the large volume of GPS data may exceed the level of TB for a metropolitan area. The storage and computation resources of traditional tools may not be enough in a short future. 

\subsection{Existing Tools}
\subsubsection{Apache Hadoop}
Apcahe Hadoop~\citep{hadoop}, first released in April 2006, is a sub-project of Lucene (a collection of industrial-strength search tools), under the umbrella of the Apache Software Foundation. Hadoop is written in Java and is a great big data tool because it can process structured and unstructured data from different sources. It parallelizes data processing utilizing computer clusters, accelerating large computations and hiding I/O latency through increased concurrency. It can also leverage its distributed file system to cheaply and reliably replicate chunks of data to nodes (computers) in the cluster, making data available locally on the machine that is processing it. Additionally, Hadoop provides to the application programmer the abstraction of map and reduce operations.

\subsubsection{Apache Pig}
Apache Pig~\citep{pig}, first released in September 2008, was initially developed by Yahoo’s researchers for executing MapReduce jobs on large datasets on a high-level of abstraction. It provides Pig Latin, a high-level scripting language, for writing data analysis code. Users write a script using the Pig Latin Language to process data in HDFS. The Pig Engine, a component of Apache Pig, converts all the scripts into a map and reduces the task for processing. However, this is a totally internal process, and thus, the programmers would not see the procedures. The result of Pig would be stored in HDFS after finishing. Compared to MapReduce, the Apache pig reduces the time of development using the multi-query approach. The same job could be completed in much less code using Pig Latin compared to using Java. A Pig Latin can be extended using user-defined functions (UDFs) that the user can write in Java, Python, JavaScript, Ruby, or Groovy and then call directly from the language.

\subsubsection{Apache Mahout}
Apache Mahout~\citep{mahout}, first released in Apr 2009, is an open-source project to provide service for the application of distributed and scalable machine learning algorithms focused on linear algebra and statistics. It is written in Java and Scala. Mahout operates in addition to Hadoop, which allows users to apply machine learning algorithms through distributed computing on Hadoop. Mahout's core algorithms include recommendation mining, clustering, classification, and frequent item-set mining. Developers are still working on Mahout but a number of algorithms have been implemented.

\subsubsection{Apache Spark}
Apache Spark~\citep{spark} started as a research project at the UC Berkeley AMPLab in 2009 and was open-sourced in early 2010. It is a cluster computing technology designed for fast computation with its own cluster management. As an extension of the Hadoop MapReduce model, it allows more types of computation including batch applications, iterative algorithms, interactive queries, and stream processing. The main feature of Spark is its in-memory cluster computing that increases the processing speed of an application by reducing the number of read/write operations to disk. Spark provides built-in APIs in Java, Scala, or Python that enable writing applications in different programming languages.

\subsubsection{Apache Kafka}
Apache Kafka~\citep{kafka}, initially released in January 2011, is an open-source distributed publish-subscribe messaging system created by LinkedIn and written in Scala and Java. It aims to provide a unified, high-throughput, and low-latency platform for handling and messaging high-volume real-time data. Kafka utilizes a binary TCP-based protocol that is optimized for efficiency and group messages to reduce the overhead of the network roundtrip. Kafka is also fault-tolerant as the messages persist on the disk and replicate within the cluster and are built on top of the ZooKeeper synchronization service.

\subsubsection{Apache Flink}
Apache Flink~\citep{flink}, initially released in May 2011, is an open-source, unified stream-processing and batch-processing framework developed by the Apache Software Foundation. The main part of Apache Flink is a distributed streaming dataflow engine that is written in Java and Scala. It executes tasks in a parallel and pipelined way and thus can perform bulk/batch processing programs, stream processing programs and iterative algorithms. In addition to providing a high-throughput, low-latency streaming engine, Flink also supports event-time processing and state management. Applications developed using Flink are fault tolerant when machines fail, as Flink relies on external fault tolerance services for maintaining configuration information and distributed synchronization.

\subsubsection{Apache Storm}
Apache Storm~\citep{storm}, initially released in September 2011, is a distributed stream processing computation framework originally created by Nathan Marz and the team at BackType. Storm was later acquired and open-sourced by Twitter and became a standard for distributed real-time processing systems in a short amount of time. Storm is written predominantly in the Clojure programming language. It uses custom created "spouts" and "bolts" to define the data manipulation process to allow batch, distributed processing of streaming data. “Spouts” would take in data and distribute to “bolts” and the functions and codes users write would be processed in bolts. The whole storm application composed of “spouts” and “bolts” could be represented by a directed acyclic graph (DAG) and form a “topology”. Data flow through edges on the graph are named streams. Together, the topology acts as a data transformation pipeline. The general topology structure is similar to a MapReduce job but data are processed in real time instead of in individual batches. Additionally, Storm topologies run indefinitely until being shut down or encounter failure, while a MapReduce job DAG must eventually end.

\subsubsection{Apache HDFS}
The Hadoop Distributed File System (HDFS)~\citep{hdfs}, first released on September 4, 2007, is a distributed file system designed to run on low-cost commodity hardware. HDFS can store and process large application datasets by storing data files across multiple machines and through parallel processing. It is highly fault tolerant, as it stores data redundantly across data nodes. HDFS relaxes a few POSIX requirements to enable streaming access to file system data. Originally built as infrastructure for the Apache Nutch web search engine project, HDFS is now an Apache Hadoop subproject.

\subsubsection{Apache HBase}
Apache HBase~\citep{hbase}, initially released in March 2008, is an open-source, column-oriented, non-relational database that evolves from Google’s Bigtable. It is written in Java and sits on top of the HDFS or Alluxio, providing Hadoop with Bigtable functionalities. It is capable of storing a large amount of sparse data in a fault-tolerant way. In addition, HBase provides compression, in-memory operation, and Bloom filters features on a per-column basis. Through the Java API and many other APIs, users can input tables from HBase into MapReduce jobs run in Hadoop, and MapReduce jobs can also output in HBase tables format. Because of the lineage with Hadoop and HDFS, HBase has been adopted worldwide. HBase cannot replace a classic SQL database, but the SQL layer and JDBC driver are provided by the Apache Phoenix project to enable the use of HBase with analytics and business intelligence applications.

\subsubsection{MongoDB}
MongoDB~\citep{MongoDB}, first released in February 2009, is an open-source document-oriented database. Classified as a NoSQL database program, MongoDB does not use regular tables and rows to store data but instead uses JSON-like documents. These documents support embedded fields and related data can be stored within them. MongoDB has optional schemas; thus, users are not required to specify the number or type of columns before inserting data. MongoDB is developed by MongoDB Inc. and licensed under the Server Side Public License (SSPL).

\subsubsection{Apache Hive}
Apache Hive~\citep{Hive}, first released in October 2010, is a data warehouse infrastructure initially developed by Facebook for querying and analyzing data. It is built on top of Apache Hadoop and provides users with an SQL-like interface for manipulating data stored in databases and file systems that integrate with Hadoop. Hive solves the problem of having to use Java API for executing SQL applications and queries over distributed data by providing its own SQL type language for querying called HiveQL or HQL. Since most data warehousing applications work with SQL-based querying languages, Hive facilitates the use of SQL-based applications to Hadoop.

\subsection{Comparison and Recommendation}
Since these tools are designed for general purposes and not dedicated for traffic data and problems, it would be necessary to compare the tools with similar functionalities and give a recommendation for choosing the suitable big data tools. The advantages and disadvantages of big data databases and tools are summarized in Table~\ref{tab:tool_comparison}. While more recent tools have new features and functionalities supported, they are also prone to potential errors. In this study, we would recommend more mature tools that are already widely used in other fields. Specifically, we would recommend the combination of Apache HDFS and Apache Hadoop for those with little or no previous experience of using big data tools and Apache Spark for those with some experience.

\begin{table}[H]
\caption{A comparison of big data tools.\label{tab:tool_comparison}}
    \begin{adjustwidth}{-\extralength}{0cm}
        \begin{tabularx}{\fulllength}{|p{3cm}|p{3.7cm}|p{5cm}|p{5cm}|}
            \toprule
            Tool & Main Purpose & Advantage & Disadvantage \\
            \midrule
            Apache Hadoop & Distributed computing & Mature and reliable. & Difficult to use. \\
            \hline
            Apache Pig & Distributed computing & High-level interface. & The need to learn Pig Latin language. \\
            \hline
            Apache Mahout & Distributed machine learning & Support for machine learning algorithms. & Performance bottleneck of default models. \\
            \hline
            Apache Spark & Distributed computing & In-memory cluster computing. Easy to use. & No automatic optimization process. \\
            \hline
            Apache Kafka & Distributed messaging & Low latency. High throughput. & Reduced performance. \\
            \hline
            Apache Flink & Distributed computing & High efficiency. Easy to use. & Immature and lack of API support. \\
            \hline
            Apache Storm & Distributed computing & Fast and fault tolerant. & Difficult to learn and use. \\
            \hline
            Apache HDFS & Database & Fault tolerant. Integrated with Apache Hadoop. & Difficult to use. \\
            \hline
            Apache HBase & Database & Real-time querying. Suitable for sparse data. Low-latency operation. & No SQL-like interface. \\
            \hline
            MongoDB & Database & Document-oriented database. & Transactions are not supported. Limited data size. High memory usage. \\
            \hline
            Apache Hive & Database & SQL-like interface. & Not a full database. No real-time querying. \\
            \bottomrule
        \end{tabularx}
    \end{adjustwidth}
\end{table}

\section{Challenges and Future Directions}
\label{sec:challenge}
While there are many successful application cases of big data for traffic estimation and prediction tasks, challenges still exist. For now, the available data sources presented in this study are still limited and can be used in statistical production only in a very limited way. There are still severe difficulties in the data generating process, which requires a joint cooperation among academia, industry and government. The legislation is another matter to consider, since many companies do not want to share but sell data or use their monopoly, \textit{e.g.}, with mobile data in their countries.

Data richness varies greatly for different transportation modes and the problems of data sparsity, high missing data ratios, data noise or lack of data still exist. Data quality, privacy and policies have not been fully considered in previous studies. For example, crowd-sourced data have the problems of low data quality, noise removal difficulty and privacy concerns. Some efforts have been made to address these challenges, e.g., sparse Bayesian learning is used for traffic state estimation with under-sampled data~\citep{babu2020sparse}. Existing data for traffic estimation and prediction tasks are heterogeneous in the spatial and temporal ranges. Cross-scale data fusion by integrating various sources is still challenging in the transportation domain.

Another challenge is the lack of ``real'' big data in the transportation domain, especially open data. While we did review many open data sources in this study, some of them have a data volume that is hard to be defined as ``big". The time range of some available datasets is not long enough for training effective deep learning models. This challenge is partially caused by the high-cost and time-consuming process of collecting some types of data. The other reason is the concern of location privacy leakage, which prohibits the collection of fine-grained data. For existing big datasets, e.g., GPS trajectories, the existing big data tools discussed in Section~\ref{sec:tools} are not yet fully exploited. With the popularity of using graph data for traffic prediction, e.g., transportation networks in graph neural networks, the existing graph processing tools cannot fully meet the computing requirements; thus, new tools are needed.

To address these challenges, some future directions are pointed out in this section.

\subsection{Heterogeneous Data Fusion}
A single data source may not be enough for traffic tasks, when different transportation modes are entangled together. Data from multiple sources can be combined to make a better estimation and prediction for traffic situations. Heterogeneous data fusion is driven by this observation, which often uses deep learning for urban big data fusion~\citep{liu2020urban}. There are already some relevant studies. For example, cellular data and loop detectors are integrated for freeway traffic speed estimation in~\cite{zhang2015accuracy}. Sparse stationary traffic sensor data are combined with GPS trace data to estimate the traffic flow in the entire road network in~\cite{gkountouna2020traffic}. License plate recognition data and cellular signaling data are combined for traffic pattern and population distribution correlation analysis in~\cite{chen2021research}. License plate recognition data and GPS trajectory data are combined for traffic flow estimation in~\cite{wang2020estimating}. Geomagnetic detector data, floating car data and license plate recognition data are combined for average link travel time extraction~\citep{guo2020reliable}. Bus transit schedule data, real-time bus location data, and cell phone data from geographical mapping software are combined to predict bus delays and a mean absolute percentage error (MAPE) of approximately 6\% is achieved~\citep{shoman2021deep}. The success of these attempts demonstrates both the necessity and the prospect of heterogeneous data fusion techniques.

\subsection{Hybrid Computing and Learning Modes}
With various data collection sources, different computing modes have been used to collect and process different types of data, e.g., cloud computing, mobile computing, edge computing, fog computing, etc. Different computing modes have different computation and communication capacities and how to integrate and utilize these computing modes with existing big data infrastructures remains challenging. Some studies are exploring in this direction. For example, based on IoT devices and fog computing, a low-cost vehicular traffic monitoring system is developed in~\cite{vergis2020low}, which collects vehicle GPS traces and uses fog devices to process the collected data and extract traffic information.

Traffic estimation and prediction problems are usually formulated as supervised problems from the perspective of machine learning. However, other learning modes can be leveraged for solving the challenges in these problems, e.g., transfer learning~\citep{pan2009survey} and generative adversarial learning~\citep{goodfellow2014generative}. Transfer learning is a potential solution for the data sparsity problem in different locations. A POI embedding mechanism is proposed in~\cite{jiang2021transfer} to fuse human mobility data and city POI data. Furthermore, CNN and LSTM are combined to capture both spatiotemporal and geographical information and mobility knowledge is transferred from one city to another, which is proven effective for improving the prediction performance for the target city with only limited data available. Existing traffic estimation and prediction methods are developed with the assumption that the traffic infrastructures remain the same. This may not hold in some cases with new urban planning. A novel off-deployment traffic estimation problem is proposed and defined in~\cite{zhang2020off} and a traffic generative adversarial network approach named TrafficGAN is further proposed to solve this problem, which is able to describe how the traffic patterns change with the travel demand and underlying road network structures.

\subsection{Distributed Solutions and Platforms}
As seen in Section~\ref{sec:tools}, big data tools are often operated in a distributed manner, or designed with the real-time parallel processing capability for streaming data, which is missing in the current store and then processes paradigms for traffic estimation and prediction studies. Some efforts for building and applying distributed solutions and platforms have been made in recent years, but mainly in the industry, e.g., Didi Brain~\footnote{\url{https://www.didiglobal.com/science/brain}} for travel demand prediction and JD Urban Spatio-Temporal Data Engine (JUST)~\footnote{\url{http://just.urban-computing.com/}} for traffic trajectory and order management. In addition to further adopting the existing tools, other new technologies may also be applied for traffic estimation, traffic prediction, or other relevant problems. For example, blockchain is used as a promising solution for traffic big data exchange trust and privacy protection~\citep{hassija2020traffic} and federated learning is used for privacy-preserving traffic flow prediction~\citep{liu2020privacy}.

\section{Conclusion}
\label{sec:conclusion}
With the development of big data, an increasing number of tools for collecting, processing, storing and utilizing data have become available. This study focuses on the tasks of traffic estimation and prediction and presents an up-to-date collection of available datasets and tools, as a reference for those who seek public resources. The collection and usage of external data are also encouraged, e.g., weather data, calendar information, emissions of air pollutants, and noise, as they provide valuable information for better estimating and predicting traffic states. While there are multiple data sources, data richness varies greatly and the data volumes are not large enough with limited spatial and temporal ranges. Off-the-shelf big data tools have not been widely used in previous studies, but there is a trend that more relevant tools would be needed with the accumulation of heterogeneous data that are beyond the abilities of traditional tools. To change this situation, challenges and future directions are pointed out, with the aim of promoting the application of big data in the transportation domain.

\authorcontributions{Conceptualization, W.J. and J.L.; methodology, W.J. and J.L.; software, W.J. and J.L.; validation, W.J. and J.L.; formal analysis, W.J. and J.L.; investigation, W.J. and J.L.; resources,W.J. and J.L.; data curation, W.J. and J.L.; writing---original draft preparation, W.J. and J.L.; writing---review and editing, W.J. and J.L.; visualization, W.J. and J.L.; supervision, W.J. and J.L.; project administration, W.J. and J.L.; funding acquisition, W.J. and J.L. All authors have read and agreed to the published version of the manuscript.}

\funding{This research received no external funding.}

\institutionalreview{Not applicable.}

\informedconsent{Not applicable.}

\dataavailability{The data would be available here: \url{https://github.com/jwwthu/DL4Traffic}.} 

\acknowledgments{Not applicable.}

\conflictsofinterest{The authors declare no conflict of interest.} 


\begin{adjustwidth}{-\extralength}{0cm}

\reftitle{References}



\bibliography{references}%

%


\end{adjustwidth}
\end{document}